\documentclass[letterpaper, 10 pt, conference]{ieeeconf}  

\IEEEoverridecommandlockouts                              

\usepackage{graphics} 
\usepackage{epsfig} 
\usepackage{mathptmx} 
\usepackage{times} 
\usepackage{amsmath} 
\usepackage{amssymb}  
\usepackage{graphicx}
\usepackage{amsmath}
\usepackage{fourier} 
\usepackage{hyperref}
\let\temp\rmdefault
\usepackage{mathpazo}
\usepackage{xcolor}
\let\rmdefault\temp
\usepackage{times}
\usepackage{booktabs}
\usepackage{xspace}
\usepackage[abs]{overpic}

\def\modelname{DENSER\xspace}

\title{\LARGE \bf \modelname: 3D Gaussians Splatting for Scene Reconstruction of Dynamic Urban Environments}

\author{Mahmud A. Mohamad, Gamal Elghazaly, Arthur Hubert, and Raphael Frank
\thanks{Mahmud A. Mohamad, Gamal Elghazaly, Arthur Hubert and Raphael Frank are with SnT - Interdisciplinary Centre for Security, Reliability and Trust, University of Luxembourg, 29 Avenue John F. Kennedy, L-1855 Luxembourg, Luxemburg. Email:
        {\tt\small \{mahmud.ali, gamal.elghazaly, arthur.hubert, raphael.frank\}@uni.lu}}%
}

\begin{document}

\maketitle
\thispagestyle{empty}
\pagestyle{empty}

\begin{abstract}
This paper presents \modelname, an efficient and effective approach leveraging 3D Gaussian \underline{s}platting (3DGS) for the \underline{r}econstruction of \underline{d}ynamic urba\underline{n} \underline{e}nvironments. While several methods for photorealistic scene representations, both implicitly using neural radiance fields (NeRF) and explicitly using 3DGS have shown promising results in scene reconstruction of relatively complex dynamic scenes, modeling the dynamic appearance of foreground objects tend to be challenging, limiting the applicability of these methods to capture subtleties and details of the scenes, especially far dynamic objects. To this end, we propose \modelname, a framework that significantly enhances the representation of dynamic objects and accurately models the appearance of dynamic objects in the driving scene. Instead of directly using Spherical Harmonics (SH) to model the appearance of dynamic objects, we introduce and integrate a new method aiming at dynamically estimating SH bases using wavelets, resulting in better representation of dynamic objects appearance in both space and time. Besides object appearance, \modelname enhances object shape representation through densification of its point cloud across multiple scene frames, resulting in faster convergence of model training. 
Extensive evaluations on KITTI dataset show that the proposed approach significantly outperforms state-of-the-art methods by a wide margin. Source codes and models will be uploaded to this repository \href{https://github.com/sntubix/denser}{\textcolor{magenta}{https://github.com/sntubix/denser}}
\end{abstract}

\section{Introduction}
Modeling dynamic 3D urban environments from images has a wide range of important applications, including building city-sale digital twins and simulation environments that can significantly reduce training and testing costs of autonomous driving systems. These applications demand efficient and high-fidelity 3D representation of the road environment from captured data and the ability to render high-quality novel views in real time. Simulation is crucial for developing and refining autonomous driving functions by providing a controlled, safe, and cost-effective testing environment. While traditional simulation tools like CARLA \cite{dosovitskiy2017carla}, LGSVL \cite{lgsvl}, and DeepDrive \cite{deepdrive} have accelerated autonomous driving development they all share of common limitation, a large \textit{sim-to-reality} gap \cite{mutsch2023model}. This gap is induced by the limitations in asset modelling and rendering that hinder model-based simulation tools their ability to fully replicate the complexities of the real world. 

To close this gap, new data-driven and photorealistic techniques based NeRFs \cite{mildenhall2020nerf} and 3DGS \cite{kerbl3Dgaussians} have shown significant capabilities for 3D scene reconstructions to visually and geometrically realistic fidelity. While NeRFs and 3DGS excel in static and small-scale scene reconstruction, reconstructing highly dynamic and complex large urban scenes remains a significant challenge. 
\begin{figure}[!t]
\centering
\begin{tabular}{cc}
\includegraphics[scale=0.09, trim=0cm 0cm 0cm 0cm,clip]{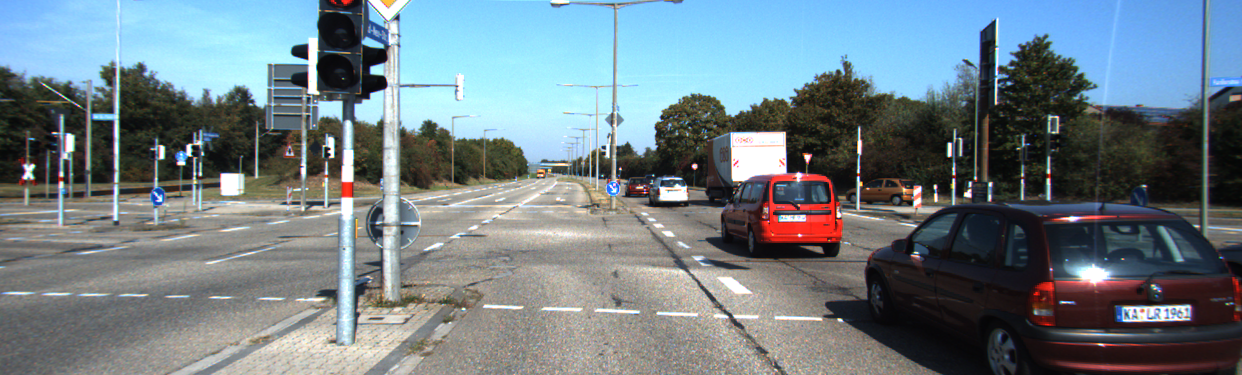} & 
\includegraphics[scale=0.09, trim=0cm 0cm 0cm 0cm,clip]{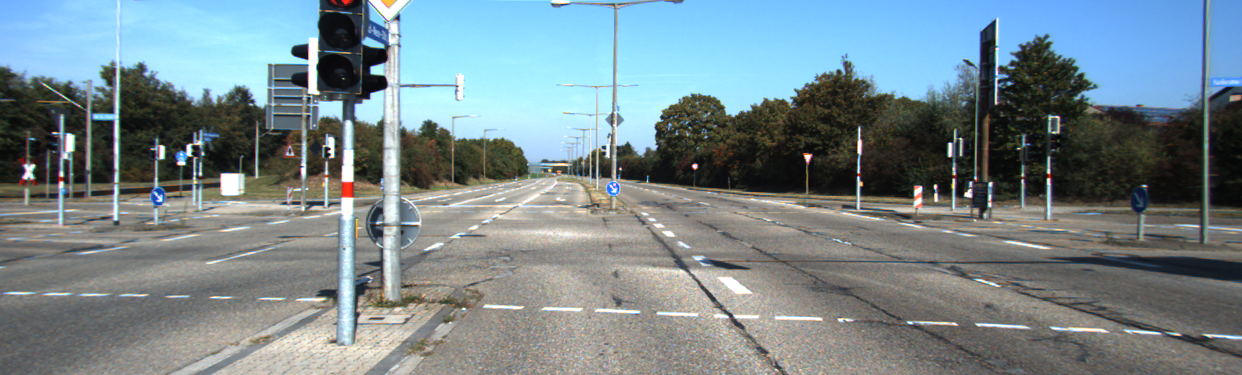} \\
\footnotesize \textbf{(a)} & \footnotesize \textbf{(b)}  \\
\includegraphics[scale=0.09]{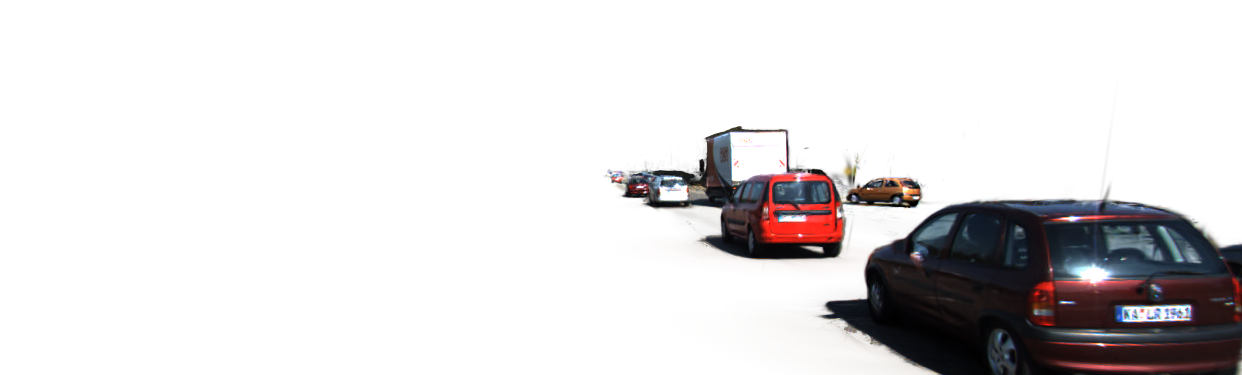} &
\includegraphics[scale=0.09]{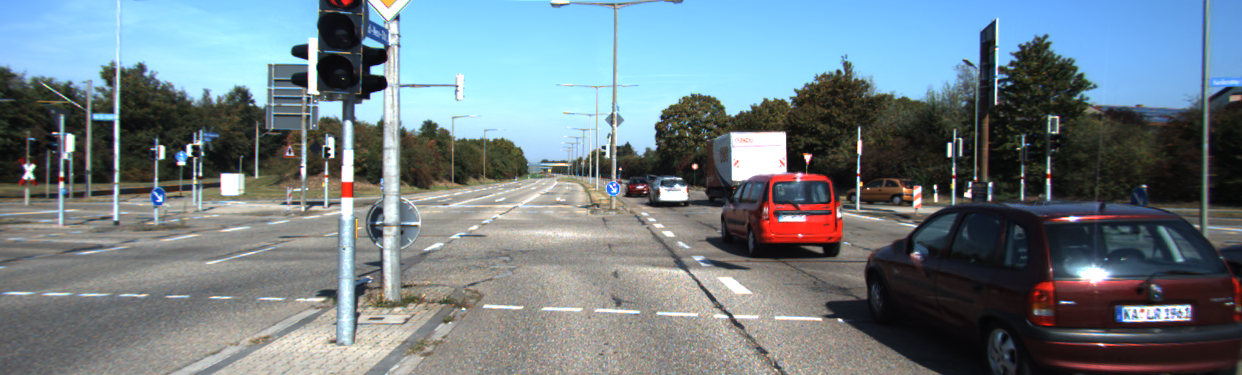} \\
 \footnotesize \textbf{(c)}  & \footnotesize \textbf{(d)}
\end{tabular}
\caption{Scene decomposition using DENSER into static background and dynamic objects and reconstruction \textbf{(a)} Ground truth
\textbf{(b)} scene decomposition: static background
\textbf{(c)} scene decomposition: dynamic objects
\textbf{(d)} scene reconstruction
}
\vspace{-0.75cm}
\label{fig:conceptdecompo}
\end{figure}

Both NeRFs \cite{mildenhall2020nerf} and 3DGS \cite{kerbl3Dgaussians} represent two distinct approaches that have shown ground-breaking scene representation results enabling photorealistic rendering and synthesising novel views of the 3D scene. While NeRFs \textit{implicitly} neural representation of the radiance field and density of the 3D scene, 3DGS \textit{explicitly} represent the scene using a large set anisotropic 3D Gaussians with associated color and opacity features. This explicit of 3DGS representation results in a faster training and rendering compared to NeRFs, thanks to parallel rasterization computed in GPUs. Despite the significant potential of of birth NeRFs and 3DGSs in static scene representation, 
their performance deteriorates considerably in dynamic scenes involving moving transient objects or when faced with changing conditions such as weather, exposure, and varying lighting \cite{dahmani2024swag, martin2021nerf}. Numerous works have already attempted to address this challenge. 
Early approaches disregarded dynamic objects and focused solely on reconstructing static components of the scene \cite{guo2023streetsurf, dahmani2024swag, martin2021nerf, tancik2022block}, rendered viewers from these approaches typically suffer from artefacts induced by transient objects. Two different approaches for dynamic scene representation have shown initial but promising results. The first represents the scene the scenes as a combination of a static and time-varying radiance fields \cite{turki2023suds, yang2023emernerf, nguyen2024rodus}. In the second approach, graph is used to represent the scene and its nodes represent static background and foreground dynamics objects, while edges maintain relationships between scene static and dynamic entities needed for scene composition over time \cite{2023mars, yan2024street, zhou2024drivinggaussian, ost2021neural}. However, most of these scene graph-based approaches do not or insufficiently consider the appearance of dynamic objects time. This paper proposes \modelname, a scene graph-based framework that significantly enhances the representation of dynamic objects and accurately models the appearance of dynamic objects in the driving scene (Fig. \ref{fig:conceptdecompo}). Instead of directly using Spherical Harmonics (SH) to model the appearance of dynamic objects, we introduce and integrate a new method aiming at dynamically estimating SH bases using wavelets, resulting in a better representation of dynamic objects appearance in both space and time. Our proposed methods achieve superior scene decomposition on the KITTI dataset.

The rest of this paper is organized as follows. Section \ref{sec:related} provides a review of related work in 3D scene reconstruction. Section \ref{sec:method} presents the proposed methodology, Section \ref{sec:exp} presents experimental results, demonstrating the effectiveness of our approach on the KITTI dataset. Finally, Section \ref{sec:conclusion} concludes this paper.

\section{Related Work}
\label{sec:related}

Dynamic scene representation has seen remarkable progress, especially in the domain of 4D neural scene representations focusing on scenes of single dynamic object, where time is considered as an additional dimension besides spatial ones \cite{attal2023hyperreel, fridovich2023k, li2021neural, lin2023high, park2021hypernerf, peng2023representing, song2023nerfplayer}. Alternative to time modulation, dynamic scenes can be modelled by coupling a deformation network to map time-varying observations to canonical deformations \cite{pumarola2021d, yang2023deformable, yang2023deformable3dgs}.
These approaches are generally limited to small-scale scenes and slight movements and are considered inadequate for complex urban environments. Furthermore, these approaches are not designed to decouple dynamic scenes into their static and dynamic primitives, e.g. instance-aware decomposition, therefore their applicability in autonomous driving simulations is limited. Alternatively, explicit decomposition of the dynamic scene facilitates accessibility and editing to manipulate these objects for simulation purposes. Scene graph has been used to model the relations between the entities composing the scene as in Neural Scene Graphs (NSG) \cite{ost2021neural}, MARS \cite{2023mars}, UniSim \cite{yang2023unisim}, StreetGaissians \cite{yan2024street}, and \cite{zhou2024drivinggaussian}. However, scene graph-based methods handle objects with limited time-varying appearances. This paper uses wavelets to enhance scene graph-based methods and how to model accurately models the appearance of dynamic objects in the driving scene.
\begin{figure*}[!ht]
\centering
\begin{overpic}[width=0.99\linewidth]{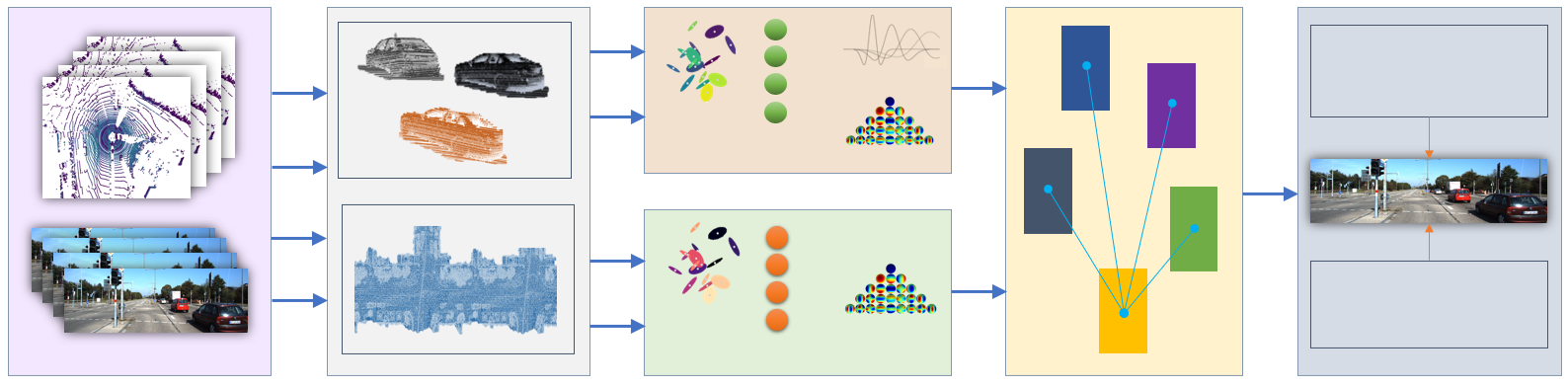}
\put(90,75){\scriptsize $I_i$}
\put(90,100){\scriptsize $T_i$}
\put(88,55){\scriptsize $O_{ij}$}
\put(90,35){\scriptsize $P_{i}$}
\put(192,48){\scriptsize $P^\mathcal{W}$}
\put(192,25){\scriptsize $I_i^b$}
\put(192,95){\scriptsize $I_i^O$}
\put(192,115){\scriptsize $P_j$}
\put(15,125){\footnotesize Scene Raw Data}
\put(120,125){\footnotesize Preprocessing}
\put(215,125){\footnotesize Dynamic Object Model}
\put(227,60){\footnotesize Background Model}
\put(335,125){\footnotesize Scene Graph}
\put(440,125){\footnotesize Rendering}
\put(217,88){\scriptsize 3DGS}
\put(217,22){\scriptsize 3DGS}
\put(259,97){\scriptsize Wavelet Basis}
\put(255,110){\scriptsize $w_i$}
\put(255,45){\scriptsize $h_{uv}^b$}
\put(270,72){\scriptsize SH Basis}
\put(270,15){\scriptsize SH Basis}
\put(210,75){\scriptsize $Y_{uv}^j = \sum_{i=1}^{d} w_i \psi_i$}
\put(423,103){\scriptsize $c = \sum_{i} c_i \alpha_i \prod_{j=1}^{i-1} (1 - \alpha_j)$}
\put(435,25){\scriptsize $\Sigma' = J W \Sigma W^T J^T$}
\put(309,38){\scriptsize $\mathcal{G}_b^\mathcal{W}$}
\put(309,105){\scriptsize $\mathcal{G}_j^\mathcal{O}$}
\put(402,70){\scriptsize $\mathcal{G}^W$}

\end{overpic}
\caption[Methodology]{\modelname Scene Composition Framework. The pipeline starts by processing raw sensor data to get a set of densified point cloud for each foreground object in its reference frame and for the static background. Object point clouds are used to initialize 3D Gaussians of dynamic objects for which wavelets are used to estimate their color appearance. Background point cloud initializes the 3D Gaussians of the static with appearance modelled using a traditional SH basis. All 3D Gaussians form a scene graph which can jointly rendered for a novel view.}
\label{fig:pipeline2}
\end{figure*}
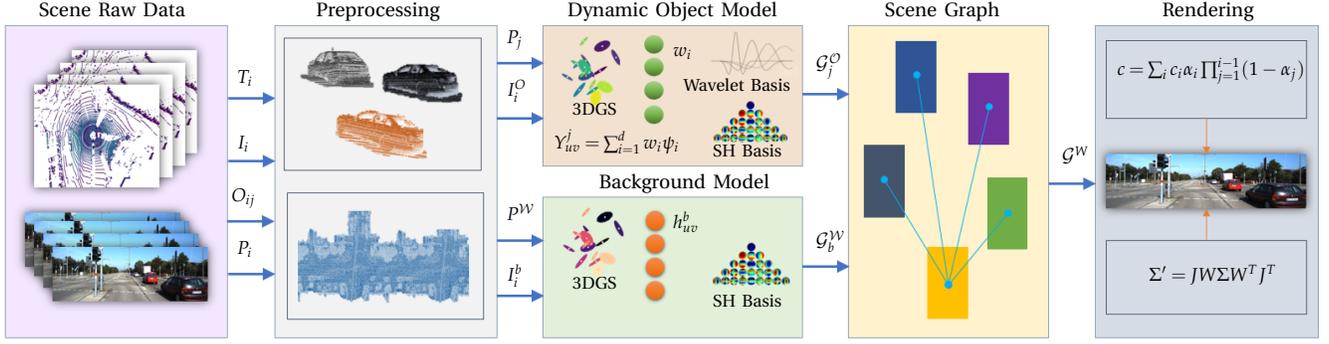

\vspace{-0.25cm}
\section{Framework and Methodology}
\label{sec:method}

\subsection{Preliminaries}
\label{sec:preliminaries}
As introduced in \cite{kerbl3Dgaussians}, 3DGS represents a scene explicitly using a finite set of 3D $n$ anisotropic Gaussians $\mathcal{G} = \{ \mathcal{G}_i\}$, each is defined by a 5-tuple $\mathcal{G}_i = \langle \mu, \, S, \, R, \, \alpha, \, c \rangle, \, \forall i=1,\, 2, \, \dots, \, n$, where $\mu \in \mathbb{R}^3$ represents its centroid, $S \in \mathbb{R}^3_{+}$ is a scale vector, $R \in SO(3)$ its rotation matrix, $\alpha \in (0, 1)$ is opacity, and $c \in \mathbb{C}^3$ is a view-dependent color, often represented using a set of coefficients in a basis of SH. The 3D volume $G_i$ occupied by the Gaussian $\mathcal{G}_i$ could be expressed as 
\begin{equation}
    G_i(x) = e^{-\frac{1}{2} (x - \mu)^T \Sigma^{-1} (x - \mu)}
\end{equation}

The covariance matrix $\Sigma$ of $\mathcal{G}_i$ could be decomposed using the rotation matrix $R$ and the scale vector $S$ as
\begin{equation}
\label{eq:sigma}
    \Sigma = R SS^T R^T
\end{equation}

For rendering, these 3D Gaussians are projected to 2D, and their covariance matrices are transformed accordingly. This involves computing a new covariance matrix \(\Sigma'\) in camera coordinates using the Jacobian of the affine approximation of the projective transformation $J$ and a viewing transformation $W$ \cite{zwicker2001ewa}
\begin{equation}
\Sigma' = J W \Sigma W^T J^T    
\end{equation}

To compute the color $c$ of a pixel is calculated using an $N$-ordered 2D splats using $\alpha$-blending
\begin{equation}
c = \sum_{i=1}^{N} c_i \alpha_i \prod_{j=1}^{i-1} (1 - \alpha_j)    
\end{equation}

While 3DGS performs well in static and object-centric small scenes, it faces challenges when dealing with scenes featuring transient objects and varying appearances \cite{}. This paper proposes a framework to model the appearance of dynamic objects by dynamically estimating the SH coefficients using wavelets, resulting in better representation of dynamic objects appearance in both space and time.

\subsection{Scene Graph Representation}

As shown in Fig.~\ref{fig:pipeline2}, the proposed framework is built on a scene graph representation accommodating both static background and dynamic objects. In \modelname, the scene is decomposed into \textit{background node} representing the static entities in the environment such roads and buildings and \textit{object nodes}, each represent a dynamic object in the scene, e.g. vehicles. Each of these nodes are represented using a set of 3D Gaussians as described in Section \ref{sec:preliminaries} that are optimized separately for each node. While the background node is directly optimized in the world reference frame $\mathcal{W}$, the object nodes are optimized in their object reference frame $\mathcal{O}_i$ that can be transformed into the world reference frame. All Gaussians corresponding to both background node and dynamic objects nodes are concatendated for rendering in a similar manner as proposed in \cite{yan2024street, kerbl3Dgaussians, zhou2024drivinggaussian}. 

Let us denote $\mathcal{G}_b^\mathcal{W}$ as the set of 3D Gaussians representing the background node and $\mathcal{G}_i^\mathcal{O}$ as the set of 3D Gaussians representing dynamic object $i$ in its object reference frame $\mathcal{O}_i$. Given the trajectory $\tau_i: \, t \to T$ of object $i$, one can extract a pose transformation matrix $T_i^\mathcal{W}(t) \in SE(3)$ representing the position and orientation of object $i$ at time $t$. Assuming the geometry of objects does not change from one pose to aonther, one can simply transform $\mathcal{G}_i^\mathcal{O}$ to the word frame by applying homogeneous transformation using $T_i^\mathcal{W}(t)$ as follows
\begin{equation}
    \mathcal{G}_i^\mathcal{W}(t) = T_i^\mathcal{W}(t) \otimes \mathcal{G}_i^\mathcal{O}
\end{equation}

The set of all Gaussians to be used for rendering can be obtained by concatenating all sets Gaussians of the static background node and transfomed dynamic objects node
\begin{equation}
    \mathcal{G}^\mathcal{W} = \bigoplus_{j=0}^{m} \mathcal{G}_j^\mathcal{W}, \,\,\,\,\,\,  \forall j=0, \, 1,\, 2, \, \dots, \, m,
\end{equation}
\noindent with $j=0$ represents the background, i.e. $\mathcal{G}_b^\mathcal{W} = \mathcal{G}_0^\mathcal{W}$ and the remaining sets of Gaussians are those of dynamic object nodes.

\subsection{Scene Decomposition}

This paper improves existing 3DGS composite scene reconstruction by enhancing the modeling of appearance of transient objects, resulting in a more realistic and consistent scene representation. The input to \modelname is a sequence $n$ frames. The frame $\mathcal{F}_i$ is defined in term of a set of $m$ tracked objects, a sensor pose $T_i$, a LIDAR point cloud $P_i$ and a set camera images $I_i$ and optionally a depth map $D_i$, $\forall i \in \{1, 2, \ldots, n\}$. Each object $j$ in the frame $i$, $O_{ij}$ is often defined by a bounding box, a tracking identifier, and an object class, $\forall j \in \{1, 2, \ldots, m\}$. Based on these inputs, \modelname starts by accumulating point clouds from over all frames in the world frame $\mathcal{W}$ while using object bounding boxes to filter the points corresponding to foreground objects. The resulting point cloud $P_b^\mathcal{W}$ is to initialize the 3D Gaussians of the background $\mathcal{G}_b^W$ for the position $\mu_b$, opacity $\alpha_b$ and covariance $\Sigma_b$ and the corresponding rotation $R_b$ and scale $S_b$ as described in (\ref{eq:sigma}) in a similar to \cite{kerbl3Dgaussians}. Besides, each Gaussian of the background is assigned a set of SH coefficients $H^b = \{ h_{uv}^b \mid 0 \leq u \leq U, -u \leq v \leq  V\}$, where $U$ and $V$ are defined by the order of SH basis defining the view-dependent color $Y_{uv}^b(\theta, \phi)$, with $\theta$ and $\phi$ define the viewing direction.
While for static scenes, the original 3DGS has shown to be capable of representing scene efficiently, it struggles to represent scenes including dynamic entities and varying appearances \cite{dahmani2024swag}. Representing the appearance of transient objects solely using SH coefficients tends to be insufficient \cite{yan2024street}. This arises mainly from the sensitivity of SH to the changes in the position of the objects in the scene and the associated changes in shadows and lighting induced by these motions. To maintain a consistent visual appearance, \modelname handles this challenge by using ($i$) densification of object point clouds across all different frames, which ensures not only a strong prior for initialization of the 3D Gaussias, but also mitigates the pose calibration errors and noisy measurements inherent in the datasets. Using the sensor pose transformation matrix $T_j$ and LIDAR point cloud $P_i$, one can apply an ROI filter defined using the bounding box of the object $O_j$ to get the point cloud $P_{ij}$ of object $j$ at frame $i$. Concatenating across all frames results in the densified point cloud $P_{j}^d$ used for initialization. ($ii$) We use a time-dependent approximation of SH bases to capture the varying appearance of dynamic objects using an orthonormal basis of wavelets with scale and translation parameters are optimizable parameters. In \modelname, the Ricker wavelet is used 
\begin{equation}
\psi(t) = \frac{2}{\sqrt{3a} \pi^{1/4}} \left(1 - \left(\frac{\tau}{a}\right)^2\right) \exp\left(-\frac{\tau^2}{2a^2}\right),
\end{equation}

\noindent where $a$ is its scale parameter and $\tau = t- b$, with $b$ is its translation parameter. The SH basis function $Y_{uv}^i(\theta, \phi)$ for object $j$ is approximated using the linear combination of child wavelets
\begin{equation}
\label{eq:Ylm}
Y_{uv}^j(t) = \sum_{i=1}^{d} w_i \psi(t, a_i, b_i)
\end{equation}
\noindent where $d$ is the dimension of the wavelet basis and $w_i$ is also an optimizable parameters. Unlike the truncated Fourier transform used in \cite{yan2024street}, wavelets are known to capture higher frequency contents even with a finite dimension of wavelet basis, resulting in significant performance to capture dynamic object details as well as varying appearances. Both ($i$) and ($ii$) constitute the genuine contribution of the present paper. 

\subsection{Optimization}

To optimize our scene, we employ a composite loss function \( \mathcal{L} \) defined as
\begin{equation}
\mathcal{L} = \mathcal{L}_{\text{color}} + \mathcal{L}_{\text{depth}} + \mathcal{L}_{\text{accum}},
\end{equation}

\noindent where \( \mathcal{L}_{\text{color}} \) represents the reconstruction loss to ensures that the predicted image \( I_{\text{pred}} \) closely matches the GT image \( I_{\text{gt}} \). This is achieved through a combination of $\mathcal{L}_1$ loss and Structural Similarity Index (SSIM) loss. The \( \mathcal{L}_{\text{1}} \) loss is given by $\mathcal{L}_{\text{1}} = \| I_{\text{gt}} - I_{\text{pred}} \|_1$
\noindent and the SSIM loss \( \mathcal{L}_{\text{SSIM}} \) is given by
$\mathcal{L}_{\text{SSIM}} = 1 - \text{SSIM}(I_{\text{gt}}, I_{\text{pred}})$
%
\noindent with \( \mathcal{L}_{\text{SSIM}} \) quantifies the similarity between two images, taking into account changes in luminance, contrast, and structure. SSIM evaluates image quality and is more sensitive to structural information. The total color loss $\mathcal{L}_{\text{color}}$ is defined in terms of $\mathcal{L}_{1}$ and $\mathcal{L}_{\text{SSIM}}$ as $\mathcal{L}_{\text{color}} = (1 - \lambda_c) \mathcal{L}_{1} + \lambda_c \mathcal{L}_{\text{SSIM}}$
%
\noindent where $\lambda_c$ is a parameter to encourage structural alignment between \(I_{\text{gt}}\) and \(I_{\text{pred}}\) \cite{kerbl3Dgaussians}. \( \mathcal{L}_{\text{depth}} \) is the mono-depth loss, which ensures that the predicted depth maps are consistent with the observed depth information. This term helps maintain the geometric consistency of the scene. The depth loss \( \mathcal{L}_{\text{depth}} \) is computed as the $\mathcal{L}_1$ loss between the predicted depth \( D_{\text{pred}} \) and the ground truth depth \(D_{\text{gt}} \) maps as $\mathcal{L}_{\text{depth}} = \lambda_d  \| D_{\text{gt}} - D_{\text{pred}} \|_1$
%
\noindent and \(\mathcal{L}_{\text{accum}} \) is the accumulation loss, which penalizes the deviation of accumulated object occupancy probabilities from the desired distributions. Specifically, it includes an entropy-based loss to ensure balanced occupancy probabilities for each object as $\mathcal{L}_{\text{accum}} = -\left( \beta \log(\beta) + (1 - \beta) \log(1 - \beta) \right)$
%
\noindent where \( \beta \) represents the object occupancy probability. 
This composite loss function facilitates the simultaneous optimization of appearance, geometry, and occupancy probabilities, ensuring a coherent and realistic reconstruction of the scene. 

\vspace{-0.5cm}
\section{Experiments and Results}
\label{sec:exp}

\subsection{Dataset and Baselines}

We conduct comprehensive evaluation of \modelname for reconstructing dynamic scenes on the KITTI dataset \cite{geiger2012we} as one of the standard benchmark for scene reconstructions in urban environments. Data frames in KITT are recorded at 10Hz. We follow the same settings and evaluation methods used in NSG \cite{ost2021neural}, MARS \cite{2023mars} and StreetGaussians \cite{yan2024street} which constitute the recent methods we use as our baseline for quantitative and qualitative comparisons.

\subsection{Implementation Details}
The training setup for our scene reconstruction utilizes the Adam optimizer across all parameters, with 30K iterations. The learning rate for the wavelets scale and translation parameters is set to \( r = 0.001 \) with \( \epsilon = 1 \times 10^{-15} \). All experiments are conducted on an NVIDIA Tesla V100-SXM2-16GB GPU. In our comparative analysis, we observed that NSG \cite{ost2021neural} and MARS \cite{2023mars} trained their models for 200K and 350K iterations, respectively, while the Street Gaussian \cite{yan2024street} reported training for 30K iterations. To determine the optimal training regimen, we tested all these configurations and found that the improvement in reconstruction quality was negligible beyond 30K iterations, with a gain of only about 0.2 in PSNR when extending from 30K to 350K iterations. Given the minimal improvement and the significant increase in training time, extending to 350K iterations was not justifiable. Specifically, training for 30K iterations takes approximately 30 minutes, whereas 350K iterations would require around 5.0 hours.

\subsection{Results and Evaluation}

\begin{table*}[t!]
    \centering
        \caption{\textnormal{Quantitative results on KITTI \cite{geiger2012we} comparing our approach with baseline methods, MARS \cite{2023mars}, SG \cite{yan2024street}, NSG \cite{ost2021neural}, and 3DGS \cite{kerbl3Dgaussians} }}
    \begin{tabular}{lcccccccccccc}
        \toprule
        \toprule
        && \multicolumn{3}{c}{KITTI - 75\%} && \multicolumn{3}{c}{KITTI - 50\%} && \multicolumn{3}{c}{KITTI - 25\%} \\
        \cmidrule{3-5} \cmidrule{7-9} \cmidrule{11-13}
        && PSNR$\uparrow$ & SSIM$\uparrow$& LPIPS$\downarrow$ && PSNR$\uparrow$ & SSIM$\uparrow$& LPIPS$\downarrow$ && PSNR$\uparrow$ & SSIM$\uparrow$& LPIPS$\downarrow$  \\
        \midrule
        3DGS \cite{kerbl3Dgaussians} && 19.19 & 0.737 & 0.172 && 19.23 & 0.739 & 0.174 && 19.06 & 0.730 & 0.180 \\
        NSG \cite{ost2021neural} && 21.53 & 0.673 & 0.254 && 21.26 & 0.659 & 0.266 && 20.00 & 0.632 & 0.281 \\
        MARS \cite{2023mars} && 24.23 & 0.845 & 0.160 && 24.00 & 0.801 & 0.164 && 23.23 & 0.756 & 0.177 \\         
        SG \cite{yan2024street} && 25.79 & 0.844 &  0.081 &&  25.52 & 0.841 & 0.084  && 24.53 & 0.824 & 0.090 \\
        Ours && \textbf{31.73} & \textbf{0.949} & \textbf{0.025} && \textbf{31.19} & \textbf{0.945} & \textbf{0.027} && \textbf{30.408} & \textbf{0.935} & \textbf{0.031} \\
        \midrule
        \bottomrule
        \end{tabular}
    \vspace{-1mm}
    \label{tab:nvs_kitti_vkitti}
\end{table*}

\begin{figure*}[t]
\centering
\begin{tabular}{c c}
\begin{tabular}{@{}c@{}}GT \end{tabular} & 
\begin{tabular}{@{}c@{}}GT \end{tabular} \\
\includegraphics[width=0.45\linewidth]{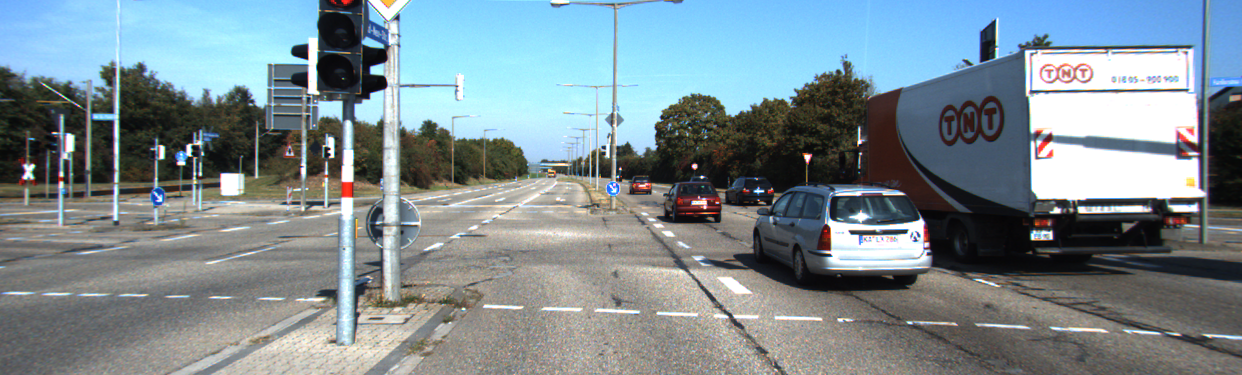} &
\includegraphics[width=0.45\linewidth]{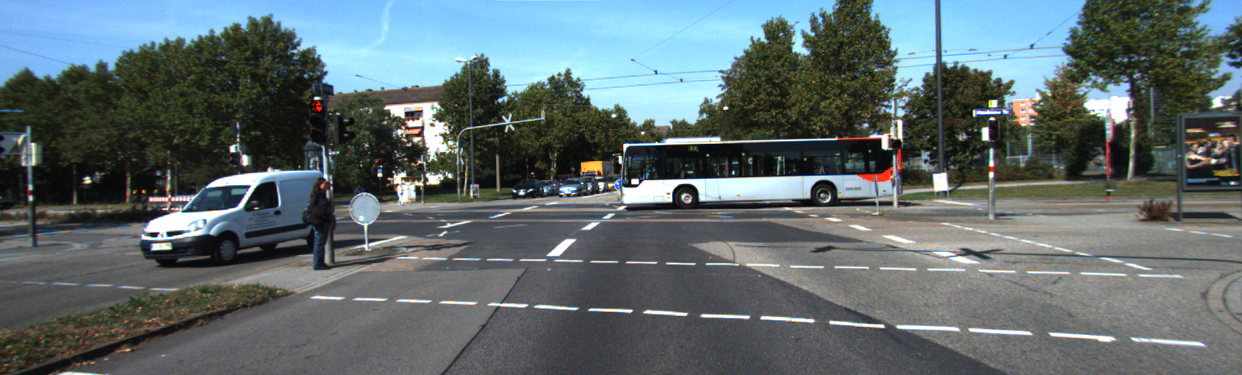} \\

\begin{tabular}{@{}c@{}}NSG \end{tabular} & 
\begin{tabular}{@{}c@{}}NSG \end{tabular} \\
\includegraphics[width=0.45\linewidth]{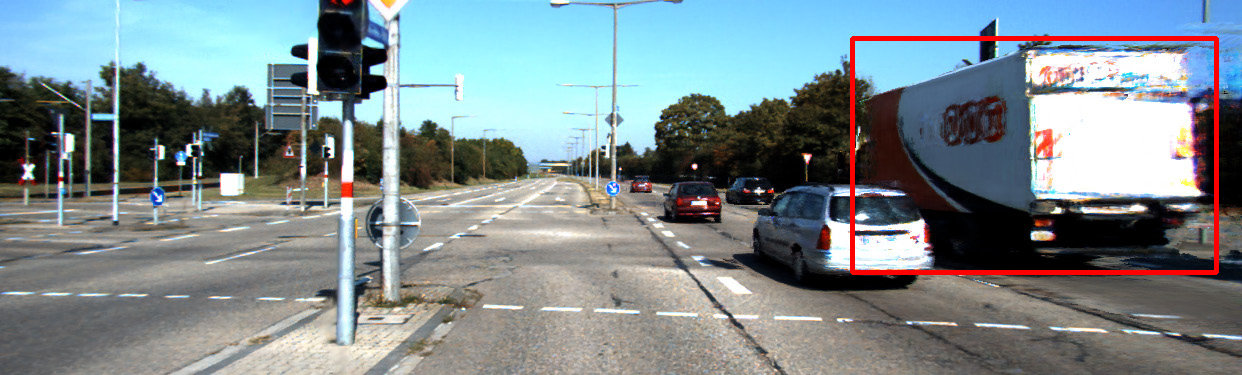} &
\includegraphics[width=0.45\linewidth]{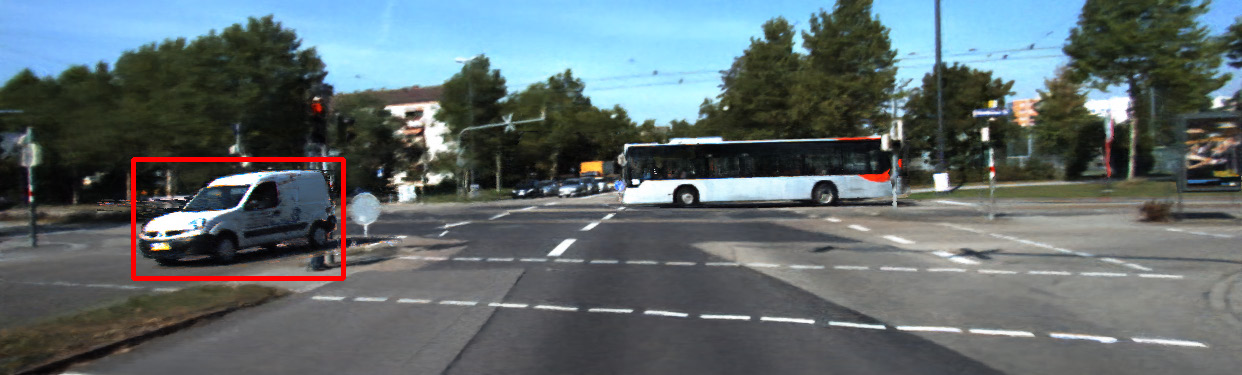} \\

\begin{tabular}{@{}c@{}}MARS \end{tabular} & 
\begin{tabular}{@{}c@{}}MARS \end{tabular} \\
\includegraphics[width=0.45\linewidth]{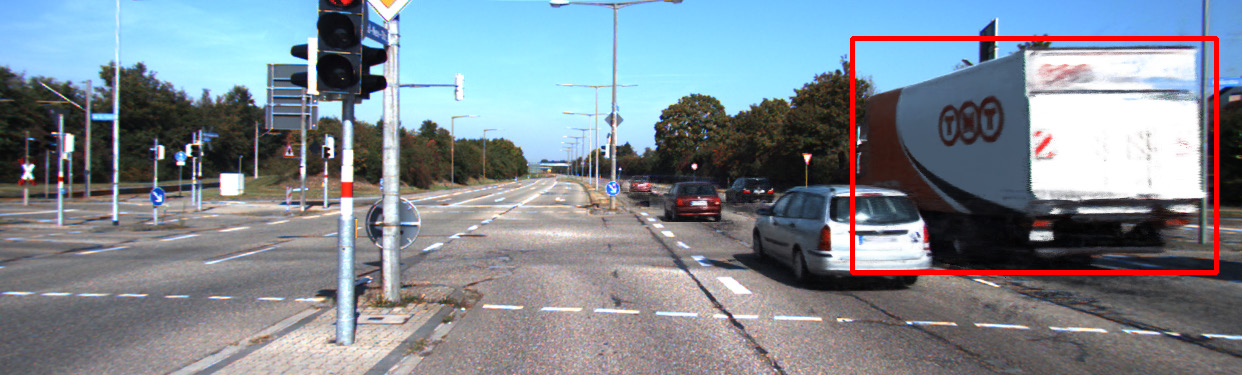} &
\includegraphics[width=0.45\linewidth]{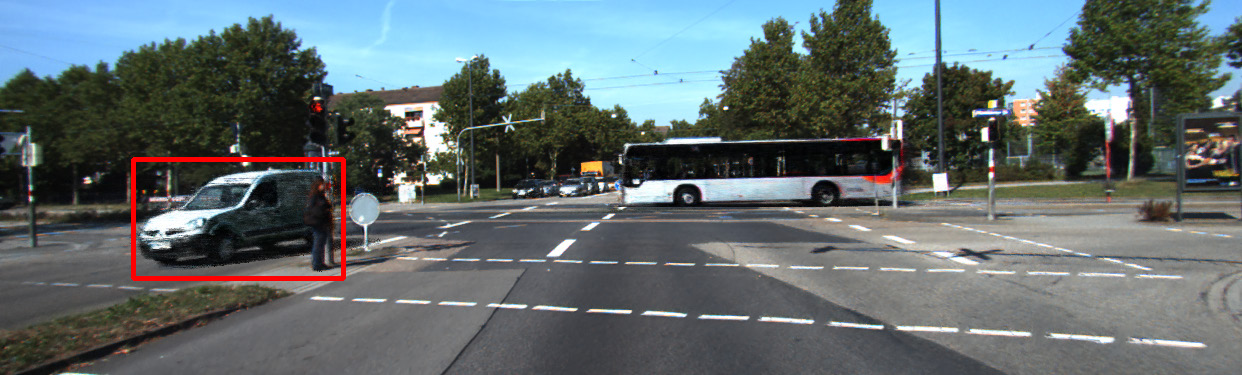} \\

\begin{tabular}{@{}c@{}}SG \end{tabular} & 
\begin{tabular}{@{}c@{}}SG \end{tabular} \\
\includegraphics[width=0.45\linewidth]{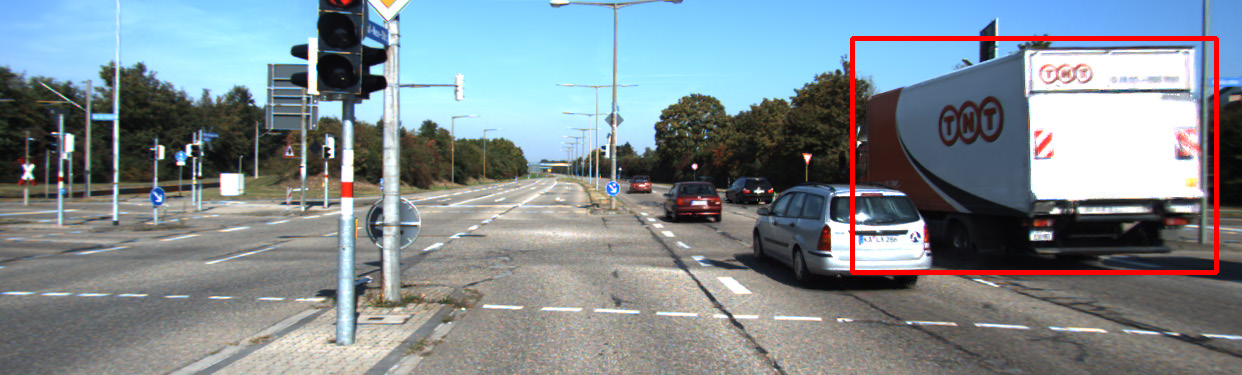} &
\includegraphics[width=0.45\linewidth]{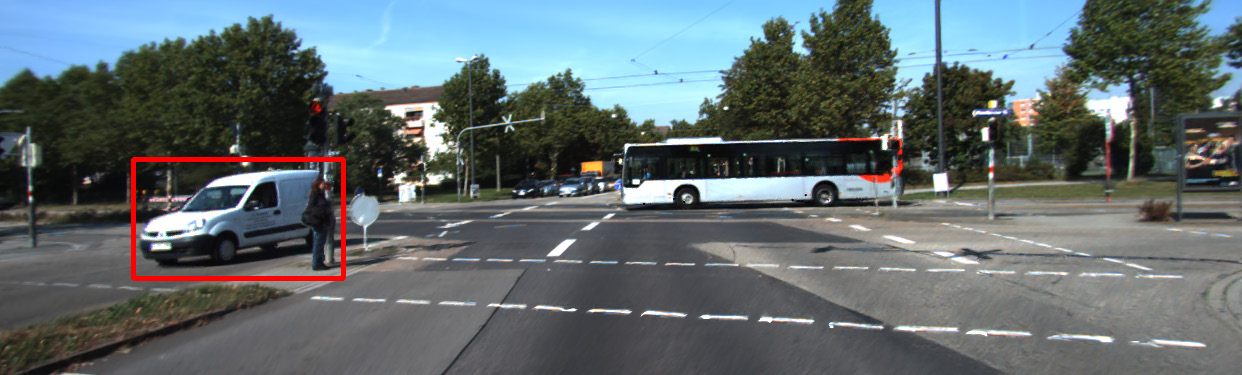} \\

\begin{tabular}{@{}c@{}}Ours \end{tabular} & 
\begin{tabular}{@{}c@{}}Ours \end{tabular} \\
\includegraphics[width=0.45\linewidth]{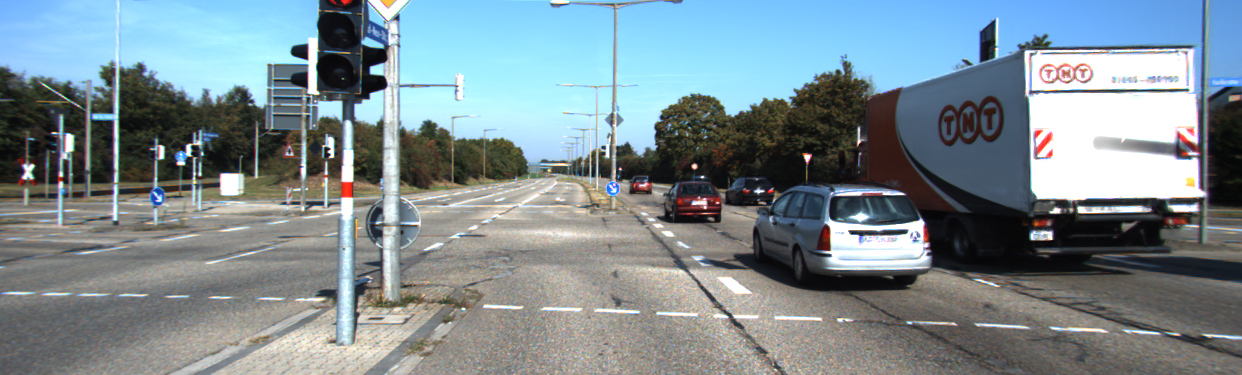} &
\includegraphics[width=0.45\linewidth]{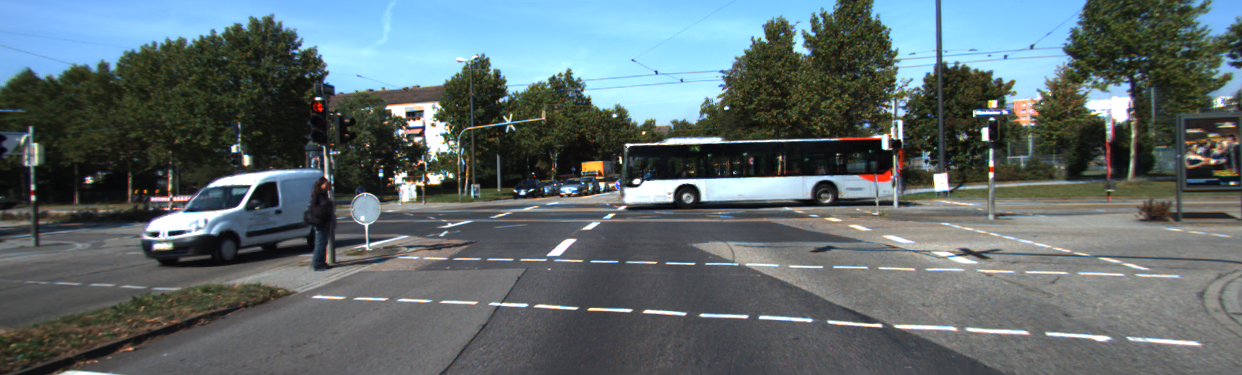} \\

\textbf{(a)} KITTI Scene 0006 & \textbf{(b)} KITTI Scene 0002
\end{tabular}
\caption{Qualitative image reconstruction comparison on KITTI dataset \cite{geiger2012we}.}
\label{fig:Name}
\end{figure*}

We conduct qualitative and quantitative comparisons against other state-of-the-art methods. These methods include, NSG \cite{ost2021neural}, which represents the background as multi-plane images and utilizes per-object learned latent codes with a shared decoder to model moving objects. MARS \cite{2023mars}, which builds the scene graph based on Nerfstudio \cite{nerfstudio}, 3D Gaussians \cite{kerbl3Dgaussians}, which models the scene with a set of anisotropic Gaussians, and StreetGaussian \cite{yan2024street}, which represents the scene as composite 3D Gaussians for foreground and background representation. We directly use the metrics reported in their respective papers to compare against our method. Table,\ref{tab:nvs_kitti_vkitti}, presents the quantitative comparison results of our method with baseline methods. As we strictly followed the same procedure and settings used in MARS and StreetGaussians (SG) to legitimately borrow their results for comparison. The rendering image resolution is 1242×375. Our approach significantly outperforms previous methods. 
The training and testing image sets in the image reconstruction setting are identical, whereas in novel view synthesis, we render frames that are not included in the training data. Specifically, we hold out one in every four frames for the 75\% split, one in every two frames for the 50\% split, and only every fourth frame is used for training in the 25\% split, resulting in 25\%, 50\%, and 75\% of the data being used for training, respectively. We adopt PSNR, SSIM, and LPIPS as metrics to evaluate rendering quality. Our model achieves the best performance across all metrics. 
Our experimental results indicate that \modelname performs exceptionally well in reconstructing dynamic scenes compared baseline methods.
The results show significant improvements in Peak Signal-to-Noise Ratio (PSNR), Structural Similarity Index (SSIM), and Learned Perceptual Image Patch Similarity (LPIPS) metrics, as detailed in Table \ref{tab:nvs_kitti_vkitti}.  
The improvements in PSNR and SSIM highlight our wavelet-based approach's effectiveness in maintaining high fidelity and structural integrity in complex environments. Furthermore, DENSER has shown to be capable of reconstructing little details, e.g. shadow at the back of truck in Scene 0006 as shown in Fig. \ref{fig:Name}, while other baseline methods are not.

\subsection{Ablation on the Dimension of Wavelet Basis}

\begin{figure}[t!]
    \centering
        \includegraphics[scale = 0.44, clip=true, trim={0.25cm 14.27cm 0cm 11.25cm}]{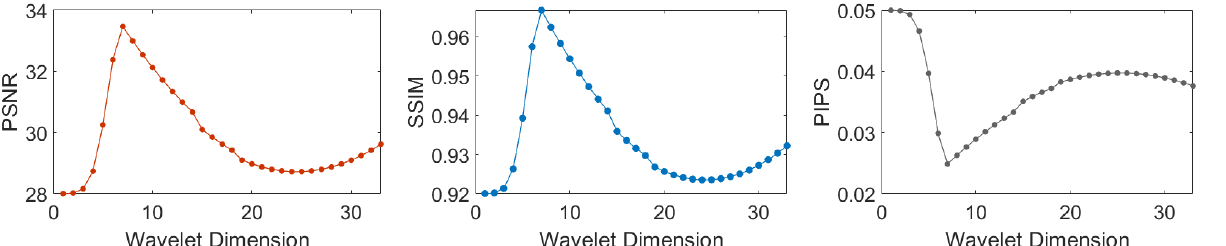} 
        \caption{Ablation: Impact of the dimension of wavelet basis on the performance of scene reconstruction}
    \label{fig:ablation}
\end{figure}

We conducted an ablation study to analyse the impact of the size of the wavelet basis, e.g. the number of wavelets used to approximate the SH functions. We run our experiments while incrementing the dimension of wavelets and analysing the impact on performance metrics (PSNR$\uparrow$, SSIM$\uparrow$ LPIPS$\downarrow$), we used for evaluation in order obtain the optimal dimension giving the best performance. As shown in Fig. \ref{fig:ablation}, the peroformance increases gradually up to $7$ wavelets and starts to degrade gradually after it. 

\subsection{Scene Editing Applications} 

\modelname enables photorealistic scene editing, such as swapping, translating, and rotating vehicles, to create diverse and realistic scenarios. This versatility allows autonomous systems to improve their performance and their ability to handle complex real-world conditions, from routine traffic to critical situations.

\subsubsection{Object Removal}
\begin{figure}[t!]
    \centering
    \begin{tabular}{cc}
        \includegraphics[width=0.475\textwidth]{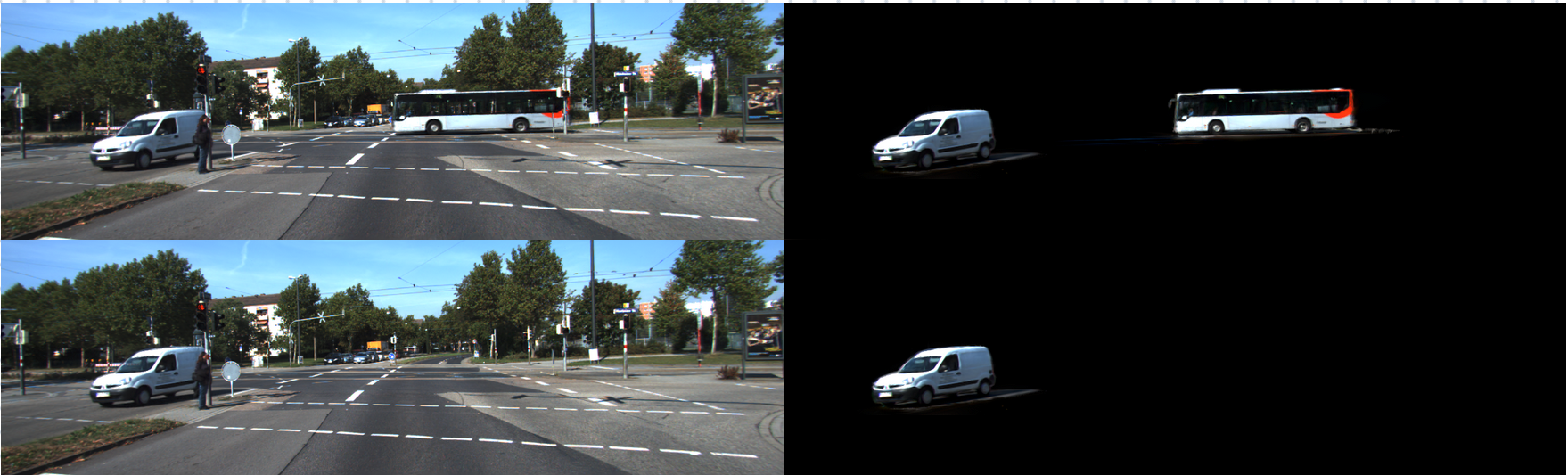}  
    \end{tabular}
            \caption[object removal]{\textbf{Object Removal:} The top row shows the GT while the bottom row displays the modified scenes where the bus have been removed.}
    \label{fig:scene_editing}
\end{figure}

To remove an object, we simply construct a deletion mask that effectively filters out the Gaussian parameters associated with the objects to be removed. The deletion mask is then applied to the Gaussian parameters of the trained model, removing the attributes associated with the unwanted objects as shown in Fig. \ref{fig:scene_editing}. 

\subsubsection{Object swapping} 
\begin{figure}[h!]
    \centering
    \begin{tabular}{cc}
        \includegraphics[width=0.475\textwidth]{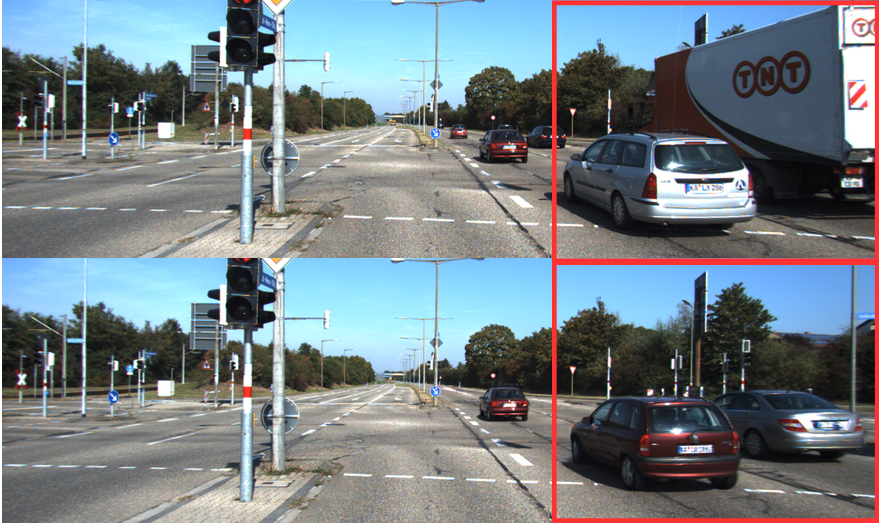} 
    \end{tabular}
            \caption[object swapping]{\textbf{Object Swapping:} The top shows the GT. In the bottom, the two vehicles within the red box of the top image have been replaced with different ones in the bottom, and some vehicles have been removed for better visualization}
    \label{fig:scene_editing_swap}
\end{figure}

Swapping vehicles within our representational framework is a straightforward process that involves a simple exchange of unique track ids associated with the two target vehicles. This manipulation results in a dynamic alteration of the scene, wherein a vehicle assumes the spatial attributes, specifically location and orientation, of the vehicle with which it has been swapped as depicted in Fig. \ref{fig:scene_editing_swap}. 

\subsubsection{Object Rotation and Translation}
\begin{figure}[t!]
    \centering
    \begin{tabular}{cc}
        \includegraphics[width=0.475\textwidth]{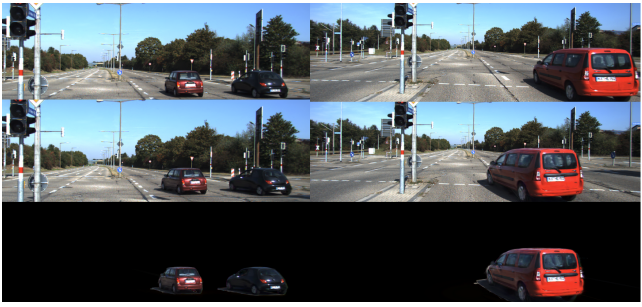} 
    \end{tabular}
            \caption[Object Rotation and Translation]{\textbf{Rotation and Translation:} The top row displays GT, illustrating the original positions and orientations of the vehicles. In the middle and bottom left images, the vehicles have been rotated. In the middle and bottom right images, the vehicle has been both rotated and translated to another lane.}
    \label{fig:trajdev}
\end{figure}

Translation and orientation modifications are implemented to adjust an object's position and heading dynamically within a 3D environment. Given an object position rotation matrix at a specific timestep \(i\), we can modify the translation and rotation to achieve desired motion maneuver. For the sake of illustration in this paper, one can shift the translation component in the plan of motion to achieve translation, while for rotation, we can transform change rotation angle about the normal to the plan of motion and calculate back the corresponding new rotation matrix to be used to replace the object as depicted in Fig. \ref{fig:trajdev}. 

\subsubsection{Trajectory Alteration}
\begin{figure}[t!]
    \centering
    \begin{tabular}{cc}
        \includegraphics[width=0.22\textwidth]{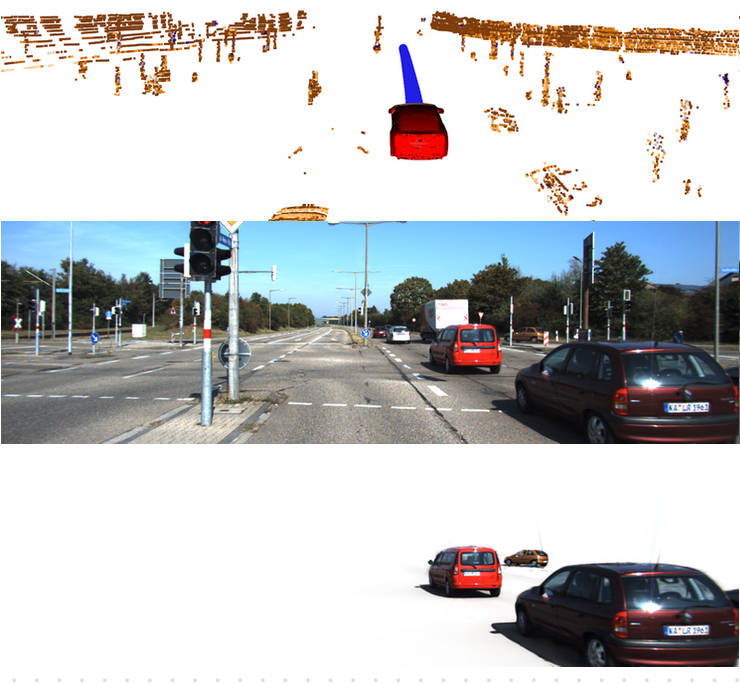} &
        \includegraphics[width=0.22\textwidth]{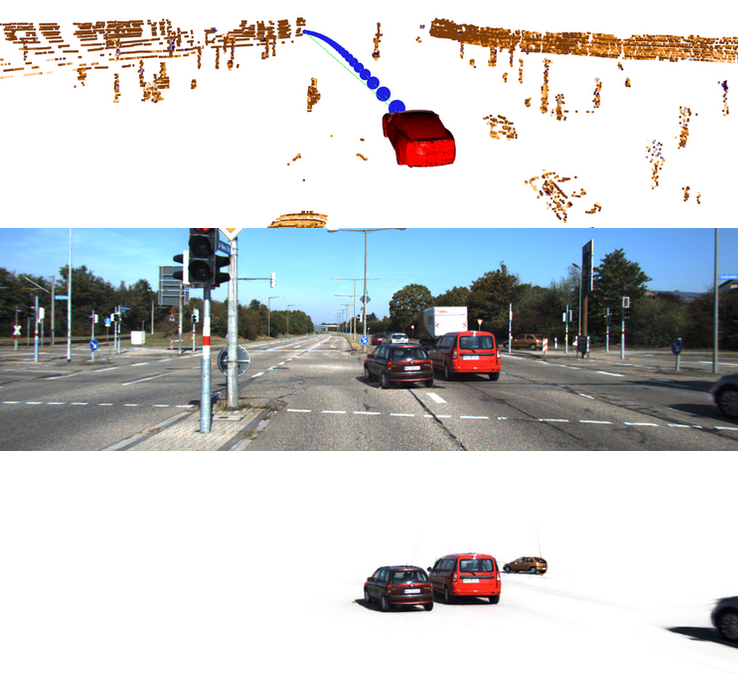} \\
    \end{tabular}
            \caption[object trajectory alteration]{\textbf{Trajectory Alteration:} The left column displays the GT trajectory and right column shows the vehicle follows a new modified path.}
    \label{fig:trajalt}
\end{figure}

A trajectory is defined as a sequence of poses. Editing the scene to allow an object follow a trajectory, one can generalize the change in rotation and translation not only between two configurations as previously illustrated but to apply this change over time to obtain a smooth change in translation and rotation as a function of time as illustrated in Fig. \ref{fig:trajalt}.

\section{Conclusion}
\label{sec:conclusion}
In this paper, we presented DENSER, a novel and efficient framework leveraging 3DGS for reconstruction of dynamic urban environments. By addressing the limitations of existing methods in modeling the appearance of dynamic objects, particularly in complex driving scenes, DENSER demonstrates significant improvements. Our approach introduces the dynamic estimation of Spherical Harmonics (SH) bases using wavelets, which enhances the representation of dynamic objects in both space and time. Furthermore, the densification of point clouds across multiple frames contributes to faster convergence during model training. Extensive evaluations on the KITTI shows that DENSER outperforms state-of-the-art techniques by a substantial margin, showcasing its effectiveness in dynamic scene reconstruction. Future work will focus on extending this approach to deformable dynamic objects such as pedestrians and cyclists.

{\small
\bibliographystyle{ieeetr}
\bibliography{bibliography}
}

\end{document}